\documentclass[11pt,a4paper]{article}
\usepackage[hyperref]{eacl2021}
\usepackage{times}
\usepackage{latexsym}

\usepackage{microtype}

\usepackage{booktabs}
\usepackage{multirow}
\usepackage{siunitx}
\usepackage{amsmath}
\usepackage{amssymb}
\usepackage{amsfonts}
\usepackage{graphicx}
\usepackage{subcaption}
\usepackage{enumitem}
\usepackage[capitalise]{cleveref}

\usepackage{etoolbox}
\robustify\bfseries

\aclfinalcopy % Uncomment this line for the final submission
\def\aclpaperid{1109} %  Enter the acl Paper ID here

\newcommand\AAST{PPT}
\newcommand\AAET{PPTX}
\newcommand\AAETLOO{\AAET{}-LOO}
\newcommand\AAETREPR{\AAET{}-REPR}
\newcommand\AAETPRAG{\AAET{}-PRAG}
\newcommand\AAETen{\AAET{}$^{EN5}$}
\newcommand\AAETPRAGsm{\AAET{}-PRAG$^S$}
\newcommand\He{\textsc{He}}
\newcommand\Ahmad{\textsc{Ahmad}}
\newcommand\Meng{\textsc{Meng}}

\title{PPT: Parsimonious Parser Transfer\\ for Unsupervised Cross-Lingual Adaptation}

\author{Kemal Kurniawan$^1$\hfill Lea Frermann$^1$\hfill Philip Schulz$^2$\thanks{~~Work
    done outside Amazon.}\hfill Trevor Cohn$^1$ \\
  $^1$School of Computing and Information Systems, University of Melbourne \\
  $^2$Amazon Research \\
  \texttt{kemal.kurniawan@student.unimelb.edu.au} \\
  \texttt{lea.frermann@unimelb.edu.au} \\
  \texttt{phschulz@amazon.com} \\
  \texttt{tcohn@unimelb.edu.au} \\}

\date{}

\begin{document}
\maketitle
\begin{abstract}
Cross-lingual transfer is a leading technique for parsing low-resource languages
in the absence of explicit supervision. Simple `direct transfer' of a learned
model based on a multilingual input encoding has provided a strong benchmark.
This paper presents a method for unsupervised cross-lingual transfer that
improves over direct transfer systems by using their output as implicit supervision as
part of self-training on unlabelled text in the target language. The method
assumes minimal resources and provides maximal flexibility by (a)~accepting any
pre-trained arc-factored dependency parser; (b)~assuming no access to source language data;
(c)~supporting both projective and non-projective parsing; and
(d)~supporting multi-source transfer. With English as the source language, we
show significant improvements over state-of-the-art transfer models on both
distant and nearby languages, despite our conceptually simpler approach. We
provide analyses of the choice of source languages for multi-source
transfer, and the advantage of non-projective parsing. Our code is available
online.\footnote{\url{https://github.com/kmkurn/ppt-eacl2021}}
\end{abstract}

\section{Introduction}

\begin{figure}
  \centering
\includegraphics[width=0.85\linewidth]{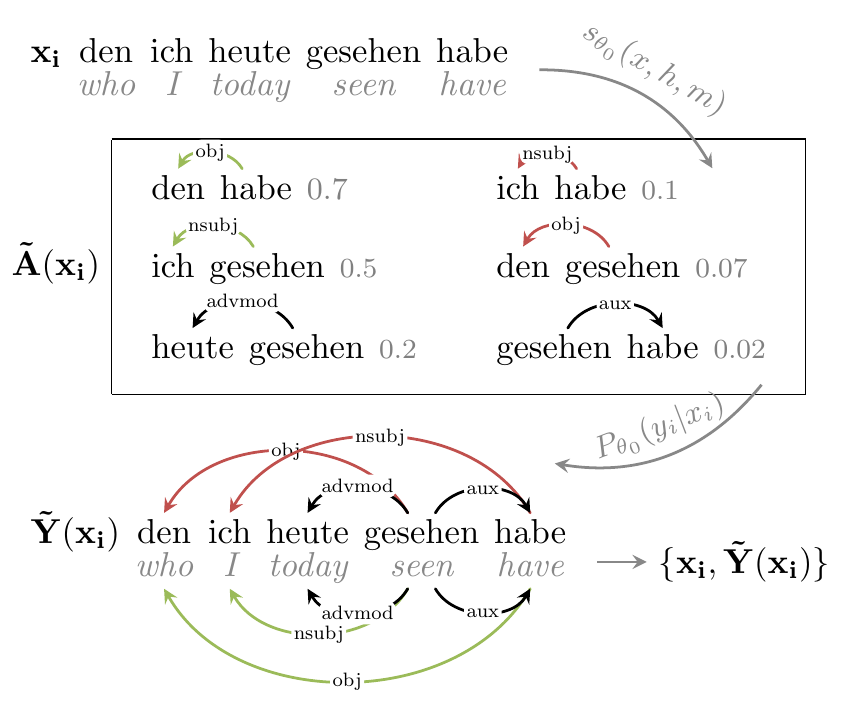}
\caption{Illustration of our technique. For a target language sentence ($x_i$),
a source parser $P_{\theta_0}$ predicts a set of candidate arcs
$\tilde{A}(x_i)$ (subset shown in the figure), and parses $\tilde{Y}(x_i)$. The
highest scoring parse is shown on the bottom (green), and the true gold parse
(unknown to the parser) on top (red). A target language parser $P_\theta$ is
then fine-tuned on a data set of ambiguously labelled sentences $\{x_i,
\tilde{Y}(x_i)\}$.}
\label{fig:illustration}
\end{figure}

Recent progress in natural language processing (NLP) has been largely driven by
increasing amounts and size of labelled datasets. The majority
of the world's languages, however, are low-resource, with little to no
labelled data available~\citep{joshi2020}. Predicting linguistic labels, such
as syntactic dependencies, underlies many downstream NLP applications, and the
most effective systems rely on labelled data. Their lack hinders the access to
NLP technology in many languages. One solution is cross-lingual model transfer,
which adapts models trained on high-resource languages to low-resource
ones. This paper presents a flexible framework for cross-lingual transfer
of syntactic dependency parsers which can leverage \emph{any} pre-trained
arc-factored dependency parser, and assumes no access to labelled target language
data.

One straightforward method of cross-lingual parsing is direct transfer. It
works by training a parser on the source language labelled data and
subsequently using it to parse the target language directly. Direct transfer is
attractive as it does not require labelled target language data,
rendering the approach fully unsupervised.\footnote{Direct transfer is also
called zero-shot transfer or   model transfer in the literature.} Recent work
has shown that it is possible to outperform direct transfer if unlabelled
data, either in the target language or a different auxiliary language, is
available~\citep{he2019a,meng2019,ahmad2019a}. Here, we focus on the
former setting and present flexible methods that can adapt a pre-trained
parser given unlabelled target data.

Despite their success in outperforming direct transfer by leveraging unlabelled
data, current approaches have several drawbacks.
First, they are limited to generative and projective parsers. However,
discriminative parsers have proven more effective, and non-projectivity is a
prevalent phenomenon across the world's languages~\citep{delhoneux2019b}.
Second, prior methods are restricted to single-source transfer, however,
transfer from multiple source languages has been shown to lead to superior
results~\citep{mcdonald2011,duong2015a,rahimi2019}. Third, they assume access to
the source language data, which may not be possible because of privacy or legal
reasons. In such source-free transfer, only a pre-trained source parser may be
provided.
%\footnote{We adopt the term source-free from SemEval-2021 Task 10
%  ``Source-Free Domain Adaptation for Semantic Processing.''
%  \url{https://machine-learning-for-medical-language.github.io/source-free-domain-adaptation/}}

We address the three shortcomings with an alternative method for unsupervised
target language adaptation~(\cref{sec:aatrn}). Our method uses high probability
edge predictions of the source parser as a supervision signal in a self-training
algorithm, thus enabling unsupervised training on the target language data. The
method is feasible for discriminative and non-projective parsing, as well as
multi-source and source-free transfer. Building on a framework introduced in
\citet{tackstrom2013}, this paper for the first time demonstrates their
effectiveness in the context of state-of-the-art neural
dependency parsers, and their generalizability across parsing frameworks. Using
English as the
source language, we evaluate on eight distant and ten nearby
languages~\citep{he2019a}.
% according to the classification given by \citet{he2019a}.
The single-source
transfer variant~(\cref{sec:aast}) outperforms previous methods by up to
\SI{11}{\percent} UAS, averaged over nearby languages. Extending the approach to
multi-source transfer~(\cref{sec:aaet}) gives further gains of \SI{2}{\percent}
UAS and closes the performance gap against the state of the art on distant
languages. In short, our contributions are:
\begin{enumerate}[noitemsep,nolistsep]
\item A conceptually simple and highly flexible framework for unsupervised target
  language adaptation, which supports multi-source and source-free transfer,
  and can be employed with any pre-trained state-of-the-art arc-factored
  parser(s);
\item Generalisation of the method of \citet{tackstrom2013} to state-of-the-art,
  non-projective dependency parsing with neural networks;
\item Up to \SI{13}{\percent} UAS improvement over state-of-the-art models,
considering
  nearby languages, and roughly equal performance over distant languages; and
\item  Analysis of the impact of choice of source
  languages on multi-source transfer quality.
\end{enumerate}

\section{Supervision via Transfer}\label{sec:aatrn}

% We present a general method for unsupervised cross-lingual parser projection,
% building on the framework of ambiguity-aware training. In ambiguity-aware
% training an input is allowed to have multiple labels to incorporate ambiguity.
% We consider the task of syntactic dependency parsing,
% which means that an input sentence $x$ may have a set of gold syntactic
% parse trees $Y$ (cf.,~\cref{fig:illustration}). During training, we want
% the parser to learn to assign high
% probabilities to these trees. Given a dataset $D=\{(x_i,Y_i)\}_{i=1}^n$, the
% ambiguity-aware training loss can be expressed as
% \begin{align}
%   \mathcal{L}(\theta)
%   &=-\frac{1}{n}\sum_{i=1}^n\log\sum_{y\in Y_i}P_\theta(y|x_i)\label{eqn:aa-loss}
% \end{align}
% where $\theta$ is the parser parameters. Note that $Y_i$ must be smaller than
% $\mathcal{Y}(x_i)$ which is the set of all trees spanning $x$ because
% $\mathcal{L}(\theta)=0$ otherwise.

In our scenario of unsupervised cross-lingual parsing, we assume the availability of a
pre-trained source parser, and unlabelled text in the target language. Thus, we aim to
leverage this data such that our cross-lingual transfer parsing method out-performs
direct transfer. One straightforward method is self-training where we
use the predictions from the source parser as supervision to train the target
parser. This method may yield decent performance as direct transfer is fairly
good to begin with. However, we may be able to do better if we also consider a set of
parse trees that have high probability under the source parser (cf.
\cref{fig:illustration} for illustration).

If we assume that the source parser can produce a set of possible trees
instead, then it is natural to use all of these
trees as supervision signal for training. Inspired by~\citet{tackstrom2013}, we formalise the method as follows.
Given an unlabelled dataset
$\{x_i\}_{i=1}^n$, the training loss can be expressed as
\begin{align}
  \mathcal{L}(\theta)
  &=-\frac{1}{n}\sum_{i=1}^n\log\sum_{y\in \tilde{Y}(x_i)}P_\theta(y|x_i)\label{eqn:aa-loss}
\end{align}
where $\theta$ is the target parser parameters and $\tilde{Y}(x_i)$ is the set
of trees produced by the source parser. Note that $\tilde{Y}(x_i)$
must be smaller than the set of all trees spanning $x$ (denoted as
$\mathcal{Y}(x_i)$ ) because $\mathcal{L}(\theta)=0$ otherwise. This training
procedure is a form of self-training, and we expect that the target parser can
learn the correct tree as it is likely to be included in $\tilde{Y}(x_i)$. Even
if this is not the case, as long as the correct arcs occur quite frequently in
$\tilde{Y}(x_i)$, we expect the parser to learn a useful signal.

We consider an arc-factored neural dependency parser where the score of a tree
is defined as the sum of the scores of its arcs, and the arc scoring function is
parameterised by a neural network. The probability of a tree is then
proportional to its score. Formally, this formulation can be expressed as
\begin{align}
  P_\theta(y|x)
  &=\frac{\exp s_\theta(x,y)}{Z(x)} \\
  s_\theta(x,y)
  &=\sum_{(h,m)\in A(y)}s_\theta(x,h,m)
\end{align}
where $Z(x)=\sum_{y\in\mathcal{Y}(x)}\exp s_\theta(x,y)$ is the partition
function, $A(y)$ is the set of head-modifier arcs in $y$, and $s_\theta(x,y)$
and $s_\theta(x,h,m)$ are the tree and arc scoring function respectively.

\subsection{Single-Source Transfer}\label{sec:aast}

% In this section, we consider the case where a single pre-trained source parser
% is provided.
% We seek to transfer this parser to a target language in which
% only unlabelled data exists. One feasible method in this scenario is
% self-traning, where we use the predictions from the source parser as the labels
% and then train a target language parser using these predicted labels. However,
% the target parser may simply mimic the source parser, thus reinforcing its
% errors. To mitigate this problem, we exploit the uncertainty of the source
% parser by predicting a set of ambiguous trees instead.

Here, we consider the case where a single pre-trained source parser
is provided and describe how the set of trees is constructed.
Concretely, for every sentence $x=w_1,w_2,\ldots,w_t$ in the target language
data, using the source parser, the set of high probability trees
$\tilde{Y}(x)$ is defined as the set of dependency trees that can be
assembled from the high probability arcs set
$\tilde{A}(x)=\bigcup_{m=1}^t\tilde{A}(x,m)$, where $\tilde{A}(x,m)$ is the set
of high probability arcs whose dependent is $w_m$. Thus, $\tilde{Y}(x)$ can be
expressed formally as
\begin{align}
  \tilde{Y}(x)
  &=\{y|y\in\mathcal{Y}(x)\wedge A(y)\subseteq\tilde{A}(x)\}.
\end{align}
$\tilde{A}(x,m)$ is constructed by adding arcs $(h,m)$ in order of decreasing
arc marginal probability until their cumulative probability exceeds a threshold
$\sigma$~\citep{tackstrom2013}. The predicted tree from the source parser is
also included in $\tilde{Y}(x)$ so the chart is never empty. This prediction is
simply the highest scoring tree. This procedure is illustrated in
\cref{fig:illustration}.

Since $\mathcal{Y}(x)$ contains an exponential number of trees, efficient
algorithms are required to compute the partition function $Z(x)$, arc marginal
probabilities, and the highest scoring tree. First, arc marginal
probabilities can be computed efficiently with dynamic programming for
projective trees~\citep{paskin2001} and Matrix-Tree Theorem for the
non-projective counterpart~\citep{koo2007,mcdonald2007,smith2007}. The same
algorithms can also be employed to compute $Z(x)$. Next, the highest scoring
tree can be obtained efficiently with Eisner's algorithm~\citep{eisner1996} or
the maximum spanning tree algorithm~\citep{mcdonald2005,chu1965,edmonds1967} for
the projective and non-projective cases, respectively.

The transfer is performed by initialising the target parser with the source
parser's parameters and then fine-tuning it with the training loss in
\cref{eqn:aa-loss} on the target language data.
Following previous works~\citep{duong2015b,he2019a}, we also regularise the
parameters towards the initial parameters to prevent them from deviating too
much since the source parser is already good to begin with. Thus, the final
fine-tuning loss becomes
\begin{align}
  \mathcal{L}^\prime(\theta)
  &=\mathcal{L}(\theta)+\lambda||\theta-\theta_0||_2^2
\end{align}
where $\theta_0$ is the initial parameters and $\lambda$ is a hyperparameter
regulating the strength of the $L_2$ regularisation.
% There are several reasons why this ambiguity-aware training may work. First, the
% target parser would not be encouraged to mimic the source parser too much as it
% has to treat all trees in $\tilde{Y}(x)$ equally. Second, although the predicted
% tree may be wrong, the correct tree is likely included in $\tilde{Y}(x)$ as the
% source parser is already good to begin with.
This single-source transfer strategy was introduced as ambiguity-aware
self-training by \citet{tackstrom2013}. A difference here is that we
regularise the target parser's parameters against the source parser's as the
initialiser, and apply the technique to modern lexicalised state-of-the-art
parsers. We refer to this transfer strategy as \AAST{} hereinafter.

Note that the whole procedure of \AAST{} can be performed even when the source
parser is trained with monolingual embeddings. Specifically, given a source
parser trained \emph{only on monolingual embeddings}, one can align pre-trained
target language word embeddings to the source embedding space using an offline
cross-lingual alignment method (e.g., of \citet{smith2017}), and use the aligned
target embeddings with the source model to compute $\tilde{Y}(x)$. Thus, our
method can be used with any pre-trained monolingual neural parser.

\subsection{Multi-Source Transfer}\label{sec:aaet}

We now consider the case where multiple pre-trained source parsers are available.
To extend \AAST{} to this multi-source case, we employ the
ensemble training method from \citet{tackstrom2013}, which we now summarise.
 We define
$\tilde{A}(x,m)=\bigcup_k\tilde{A}_k(x,m)$ where $\tilde{A}_k(x,m)$ is the set
of high probability arcs obtained with the $k$-th source parser. The rest of the
procedure is exactly the same as \AAST{}. Note that we need to select one source
parser as the main source to initialise the target parser's parameters with.
%This multi-source transfer strategy was also introduced by \citet{tackstrom2013}
%as ambiguity-aware ensemble training.
Henceforth, we refer to this method
as \AAET{}.

Multiple source parsers may help transfer better because each parser will encode
different syntactic biases from the languages they are trained on. Thus, it is
more likely for one of those biases to match that of the target language instead
of using just a single source parser. However, multi-source transfer may also
hurt performance if the languages have very different syntax, or the source parsers
are of poor quality, which can arise from poor quality cross-lingual word embeddings.
% or if the languages are very different?

\section{Experiments}

\subsection{Setup}\label{sec:setup}

We run our experiments on Universal Dependency Treebanks v2.2~\citep{nivre2018}.
We reimplement the self-attention graph-based parser of \citet{ahmad2019} that
has been used with success for cross-lingual dependency parsing. Averaged over 5
runs, our reimplementation achieves \SI{88.8}{\percent} unlabelled attachment
score (UAS) on English Web Treebank using the same
hyperparameters,\footnote{Reported in \cref{tbl:src-hyper}.} slightly below
their reported \SI{90.3}{\percent} result.\footnote{UAS and LAS are reported excluding punctuation tokens.}
  We select the run with the highest labelled attachment score
(LAS) as the source parser. We obtain cross-lingual word embeddings with the
offline transformation of \citet{smith2017} applied to fastText pre-trained word
vectors~\citep{bojanowski2017}. We include the universal POS tags as inputs by
concatenating the embeddings with the word embeddings in the input layer. We
acknowledge that the inclusion of gold POS tags does not reflect a realistic
low-resource setting where gold tags are not available, which we discuss more in
\cref{sec:results}. We evaluate on 18 target languages that are divided into two
groups, distant and nearby languages, based on their distance from English as defined by
\citet{he2019a}.\footnote{We exclude Japanese and Chinese based on \citet{ahmad2019},
  who reported atypically low performance on these two languages, which they attributed to the low quality of their cross-lingual
  word embeddings. In subsequent work they excluded these languages
  \citep{ahmad2019a}.}

During the
unsupervised fine-tuning, we compute the training loss over
all trees regardless of projectivity (i.e. we use Matrix-Tree Theorem to compute
\cref{eqn:aa-loss}) and discard sentences longer than 30 tokens to avoid
out-of-memory error. Following \citet{he2019a}, we fine-tune on the target
language data for 5 epochs, tune the hyperparameters (learning rate and
$\lambda$) on Arabic and Spanish using LAS, and use these
values\footnote{Reported in \cref{tbl:main-hyper}.} for the distant and
nearby languages, respectively. We set the threshold $\sigma=0.95$ for both \AAST{} and \AAET{} following
\citet{tackstrom2013}. We keep the rest of the hyperparameters (e.g., batch
size) equal to those of \citet{ahmad2019}. For \AAET{}, unless otherwise stated, we
consider a leave-one-out scenario where we use all languages except the target
as the source language. We use the same hyperparameters as the English parser to
train these non-English source parsers and set the English parser as the main
source.

\subsection{Comparisons}

We compare \AAST{} and \AAET{} against several recent unsupervised transfer systems.
First, \He{} is a neural lexicalised DMV parser with normalising flow that uses
a language modelling objective when fine-tuning on the unlabelled target
language data~\citep{he2019a}. Second, \Ahmad{} is an adversarial training
method that attempts to learn language-agnostic
representations~\citep{ahmad2019a}. Lastly, \Meng{} is a constrained inference
method that derives constraints from the target corpus statistics to aid
inference~\citep{meng2019}. We also compare against direct transfer (DT) and
self-training (ST) as our baseline systems.\footnote{ST requires significantly
  less memory so we only discard sentences longer than 60 tokens. Complete
  hyperparameter values are shown in \cref{tbl:main-hyper}.}

\subsection{Results}\label{sec:results}

\begin{table*}\small
  \centering
  \sisetup{
    table-format = 2.1,
    table-auto-round
  }
  \begin{tabular}{@{}c*{12}{S}@{}}
    \toprule
    \multirow{2}{*}{Target} & \multicolumn{7}{c}{UAS} & \multicolumn{5}{c}{LAS} \\
    \cmidrule(lr){2-8} \cmidrule(l){9-13}
    & {DT} & {ST} & {\AAST{}} & {\AAET{}} & {\He{}} & {\Ahmad{}} & {\Meng{}} & {DT} & {ST} & {\AAST{}} & {\AAET{}} & {\Ahmad{}} \\
    \midrule
    fa & 37.53 & 38.0 & 39.47 & 53.58 & \bfseries 63.20 & {---} & {---} & 29.24 & 30.5 & 31.60 & \bfseries 44.52 & {---} \\
    ar$^\dag$ & 37.60 & 39.2 & 39.46 & 48.29 & \bfseries 55.44 & 38.98 & 47.3 & 27.34 & 30.0 & 29.94 & \bfseries 38.48 & 27.89 \\
    id & 51.63 & 49.9 & 50.28 & \bfseries 71.91 & 64.20 & 51.57 & 53.1 & 45.23 & 44.4 & 44.72 & \bfseries 59.02 & 45.31 \\
    ko & 35.07 & 37.1 & \bfseries 37.47 & 34.59 & 37.03 & 34.23 & 37.1 & 16.57 & \bfseries 18.2 & 17.99 & 16.11 & 16.08 \\
    tr & 36.93 & 38.1 & \bfseries 39.16 & 38.44 & 36.05 & {---} & 35.2 & 18.49 & 19.5 & 19.04 & \bfseries 20.60 & {---} \\
    hi & 33.70 & 34.7 & 33.96 & 36.39 & 33.17 & 37.37 & \bfseries 52.4 & 25.40 & 26.6 & 26.37 & \bfseries 28.27 & 28.01 \\
    hr & 61.98 & 63.4 & 63.79 & \bfseries 71.90 & 65.31 & 63.11 & 63.7 & 51.87 & 54.2 & 54.18 & \bfseries 61.17 & 53.62 \\
    he & 56.62 & 59.2 & 60.49 & 64.17 & \bfseries 64.80 & 57.15 & 58.8 & 47.61 & 50.5 & 51.11 & \bfseries 53.92 & 49.36 \\
    \addlinespace
    average & 43.88 & 44.95 & 45.51 & \bfseries 52.41 & \bfseries 52.40 & {---} & {---} & 32.72 & 34.24 & 34.37 & \bfseries 40.26 & {---} \\
    \cmidrule(r){1-8} \cmidrule(l){9-13}
    bg & 77.68 & 80.0 & 81.22 & \bfseries 81.92 & 73.57 & 79.72 & 79.7 & 66.22 & 68.9 & 70.02 & \bfseries 70.24 & 68.39 \\
    it & 77.89 & 79.7 & 81.36 & \bfseries 83.65 & 70.68 & 80.70 & 82.0 & 71.07 & 74.0 & 75.53 & \bfseries 77.73 & 75.57 \\
    pt & 74.07 & 76.3 & 77.07 & \bfseries 80.95 & 66.61 & 77.09 & 77.5 & 65.05 & 67.6 & 68.26 & \bfseries 70.64 & 67.81 \\
    fr & 74.80 & 77.5 & 78.64 & \bfseries 80.57 & 67.66 & 78.31 & 79.1 & 68.13 & 71.7 & 72.76 & \bfseries 74.46 & 73.29 \\
    es$^\dag$ & 72.45 & 74.9 & 75.21 & \bfseries 78.25 & 64.28 & 74.08 & 75.8 & 63.78 & 66.5 & 67.01 & \bfseries 69.15 & 65.84 \\
    no & 77.86 & 80.4 & \bfseries 81.21 & 80.01 & 65.29 & 80.98 & 80.4 & 69.06 & 71.9 & 72.66 & 71.75 & \bfseries 73.10 \\
    da & 75.33 & 76.0 & \bfseries 77.29 & 76.57 & 61.08 & 76.25 & 76.6 & 66.25 & 67.4 & \bfseries 68.55 & 67.86 & 68.03 \\
    sv & 78.89 & 80.5 & \bfseries 82.09 & 81.03 & 64.43 & 80.43 & 80.5 & 71.12 & 72.7 & 74.20 & 72.72 & \bfseries 76.68 \\
    nl & 68.00 & 68.9 & 69.85 & \bfseries 74.39 & 61.72 & 69.23 & 67.6 & 59.47 & 60.7 & 61.53 & \bfseries 65.39 & 60.51 \\
    de & 66.79 & 69.9 & 69.52 & \bfseries 74.05 & 69.52 & 71.05 & 70.8 & 56.40 & 60.0 & 59.69 & \bfseries 63.45 & 61.84 \\
    \addlinespace
    average & 74.38 & 76.41 & 77.35 & \bfseries 79.14 & 66.48 & 76.78 & 77.0 & 65.66 & 68.14 & 69.02 & \bfseries 70.34 & 69.11 \\
    \bottomrule
  \end{tabular}
  \caption{Test UAS and LAS (avg. 5 runs) on distant (top) and nearby (bottom)
    languages, sorted from most distant (fa) to closest (de) to English. \AAET{} is
    trained in a leave-one-out fashion. The numbers for \He{}, \Ahmad{}, and
    \Meng{} are obtained from the corresponding papers, direct transfer (DT) and
    self-training (ST) are based on our own implementation. \dag~indicates languages used for
    hyper-parameter tuning, and thus have additional supervision through the use
    of a labelled development set.}\label{tbl:main}
\end{table*}

\cref{tbl:main} shows the main results. We observe that fine-tuning via
self-training already helps DT, and by incorporating multiple high probability
trees with \AAST{}, we can push the performance slightly higher on most
languages, especially the nearby ones. Although not shown in the table, we
also find the \AAST{} has up to 6x lower standard deviation than ST, which makes \AAST{}
preferrable to ST. Thus, we exclude ST as a baseline from our subsequent
experiments. Our results seem to agree with that of
\citet{tackstrom2013} and suggest that \AAST{} can also be employed for neural
parsers. Therefore, it should be considered for target language adaptation if unlabelled
target data is available. Comparing to \He{}~\citep{he2019a}, \AAST{} performs worse
on distant languages, but better on nearby languages. This finding means that if
the target language has a closely related high-resource language, it may be
better to transfer from that language as the source and use \AAST{} for adaptation.
Against \Ahmad{}~\citep{ahmad2019a}, \AAST{} performs better on 4 out of 6 distant
languages. On nearby languages, the average UAS of \AAST{} is higher, and the
average LAS is on par. This result shows that leveraging unlabelled data for
cross-lingual parsing without access to the source data is feasible. \AAST{} also
performs better than \Meng{}~\citep{meng2019} on 4 out of 7 distant languages, and
slightly better on average on nearby languages. This finding shows that \AAST{}
is competitive to their constrained inference method.

Also reported in \cref{tbl:main} are the ensemble results for \AAET{}, which are particularly strong.
\AAET{} outperforms \AAST{}, especially
on distant languages with the average UAS and LAS absolute improvements of
\SI{7}{\percent} and \SI{6}{\percent} respectively. This finding suggests that
\AAET{} is indeed an effective method for multi-source transfer of neural
dependency parsers. It also gives further evidence that multi-source transfer is
better than the single-source counterpart. \AAET{} also closes the gap against the
state-of-the-art adaptation of \citet{he2019a} in terms of average UAS on
distant languages. This result suggests that \AAET{} can be an option for languages
that do not have a closely related high-resource language to transfer from.

\paragraph{Treebank Leakage}

\begin{figure}
  \centering
  \includegraphics[width=0.82\linewidth]{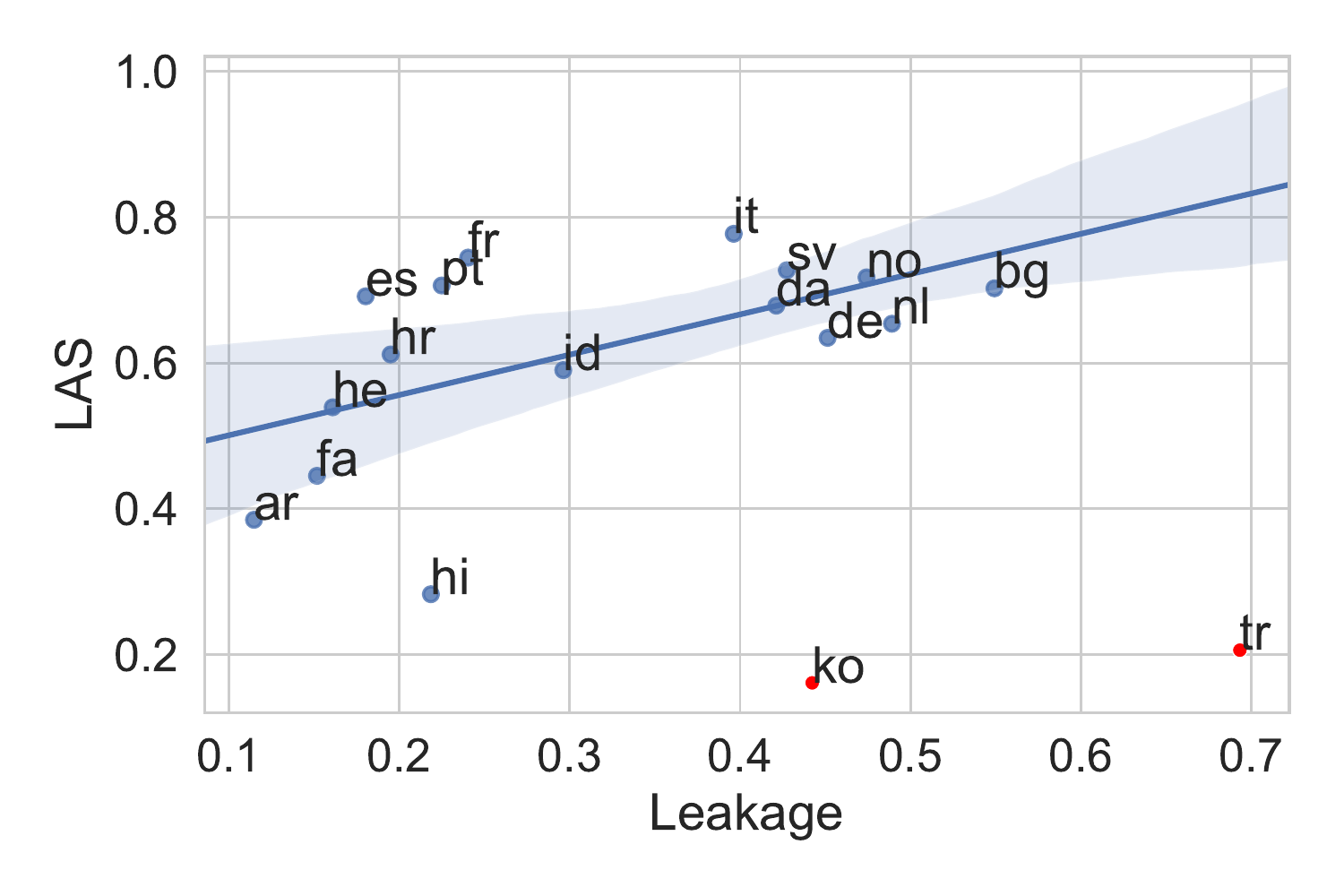}
  \caption{Relationship between treebank leakage and LAS for \AAET{}. Shaded
    area shows \SI{95}{\percent} confidence interval. Korean and Turkish (in
    red) are excluded when computing the regression line.}\label{fig:leakage}
\end{figure}

The success of our cross-lingual transfer can be attributed in part to treebank
leakage, which measures the fraction of dependency trees in the test set that are
isomorphic to a tree in the training set (with potentially different words);
accordingly these trees are not entirely unseen. Such leakage
has been found to be a particularly strong predictor for parsing
performance in monolingual parsing~\citep{sogaard2020}. \cref{fig:leakage} shows
the relationship between treebank leakage and parsing accuracy, where the
leakage is computed between the English training set as source and the target
language's test set. Excluding outliers which are Korean and Turkish because of
their low parsing accuracy despite the relatively high leakage, we find that there is a fairly strong
positive correlation ($r=0.57$) between the amount of leakage and accuracy.
The same trend occurs with DT, ST, and \AAST{}. This finding suggests that
cross-lingual parsing is also affected by treebank leakage just like monolingual
parsing is, which may present an opportunity to find good sources for transfer.

\paragraph{Use of Gold POS Tags}

As we explained in \cref{sec:setup}, we restrict our experiments to gold POS
tags for comparison with prior work. However, the use of gold POS tags does not
reflect a realistic low-resource setting where one may have to resort to
automatically predicted POS tags. \citet{tiedemann2015} has shown that
cross-lingual delexicalised parsing performance degrades when predicted POS tags
are used. The degradation ranges from 2.9 to 8.4 LAS points depending on the
target language. Thus, our reported numbers in \cref{tbl:main} are likely to
decrease as well if predicted tags are used, although we expect the decline is
not as sharp because our parser is lexicalised.

\subsection{Parsimonious Selection of Sources for \AAET{}}

\begin{figure*}
  \centering
  \includegraphics[width=\linewidth]{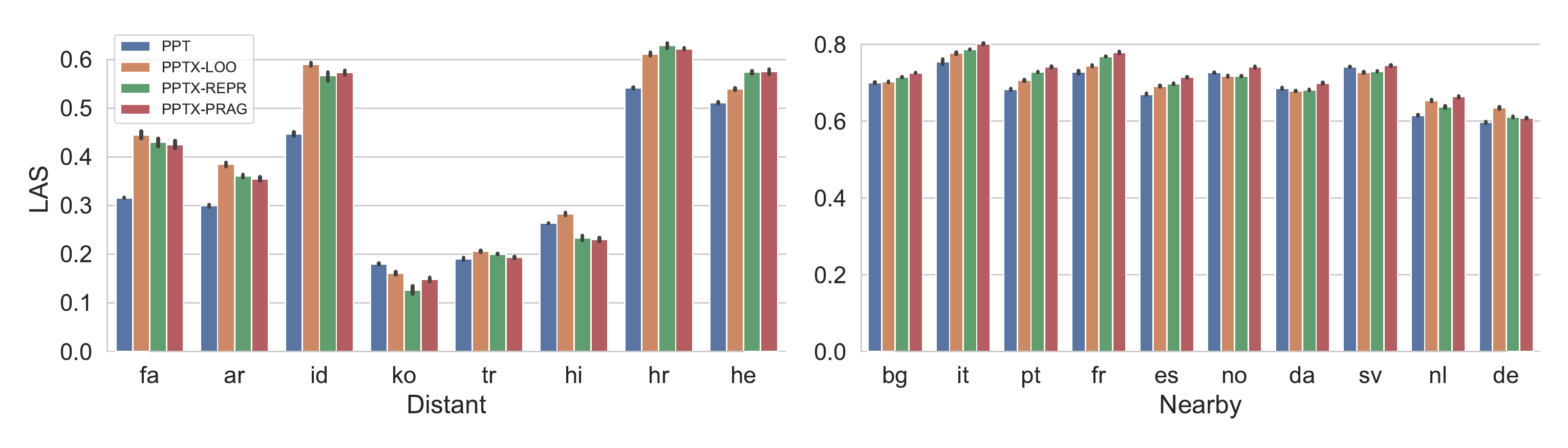}
  \caption{Comparison of selection of source languages for \AAET{} on distant and
    nearby languages, sorted from most distant (fa) to closest (de) to English.
    \AAETLOO{} is trained in a leave-one-out fashion. \AAETREPR{} uses the
    representative source language set, while \AAETPRAG{} is adapted from five
    high-resource languages. A source language is excluded from the source if it
    is also the target language.}\label{fig:src-select}
\end{figure*}

In our main experiment, we use all available languages as source for \AAET{} in a
leave-one-out setting. Such a setting may be justified to cover as many
syntactic
biases as possible, however, training dozens of parses may be impractical. In
this experiment, we consider the case where we can
train only a handful of source parsers. We investigate two selections of
source languages: (1) a representative selection (\AAETREPR{}) which covers as
many language families as possible and (2) a pragmatic selection (\AAETPRAG{})
containing truly high-resource languages for which quality pre-trained parsers
are likely to exist. We restrict the selections to 5 languages each. For
\AAETREPR{}, we use English, Spanish, Arabic, Indonesian, and Korean as source
languages. This selection covers Indo-European (Germanic and Romance),
Afro-Asiatic, Austronesian, and Koreanic language families respectively. We use
English, Spanish, Arabic, French, and German as source languages for \AAETPRAG{}.
The five languages are classified as exemplary high-resource languages by
\citet{joshi2020}. We exclude a language from the source if it is also the
target language, in which case there will be only 4 source languages. Other than
that, the setup is the same as that of our main
experiment.\footnote{Hyperparameters are tuned; values are shown in
  \cref{tbl:main-hyper}.}

We present the result in \cref{fig:src-select} where we also include the results for
\AAST{}, and \AAET{} with the leave-one-out setting (\AAETLOO{}). We report only LAS
since UAS shows a similar trend. We observe that both \AAETREPR{} and
\AAETPRAG{} outperform \AAST{} overall. Furthermore, on nearby
languages except Dutch and German, both \AAETREPR{} and \AAETPRAG{}
outperform \AAETLOO{}, and \AAETPRAG{} does best overall. In contrast,
no systematic difference between the three \AAET{} variants emerges on
distant languages. This finding suggests that instead of
training dozens of source parsers for \AAET{}, training just a handful of them is
sufficient, and a ``pragmatic'' selection of a small number of high-resource
source languages seems to
be an efficient strategy. Since pre-trained parsers for these
languages are most
likely available, it comes with the additional advantage of alleviating the need
to train parsers at all, which makes our
method even more practical.

\paragraph{Analysis on Dependency Labels}

\begin{figure}[!h]
  \centering
  \includegraphics[width=0.9\linewidth]{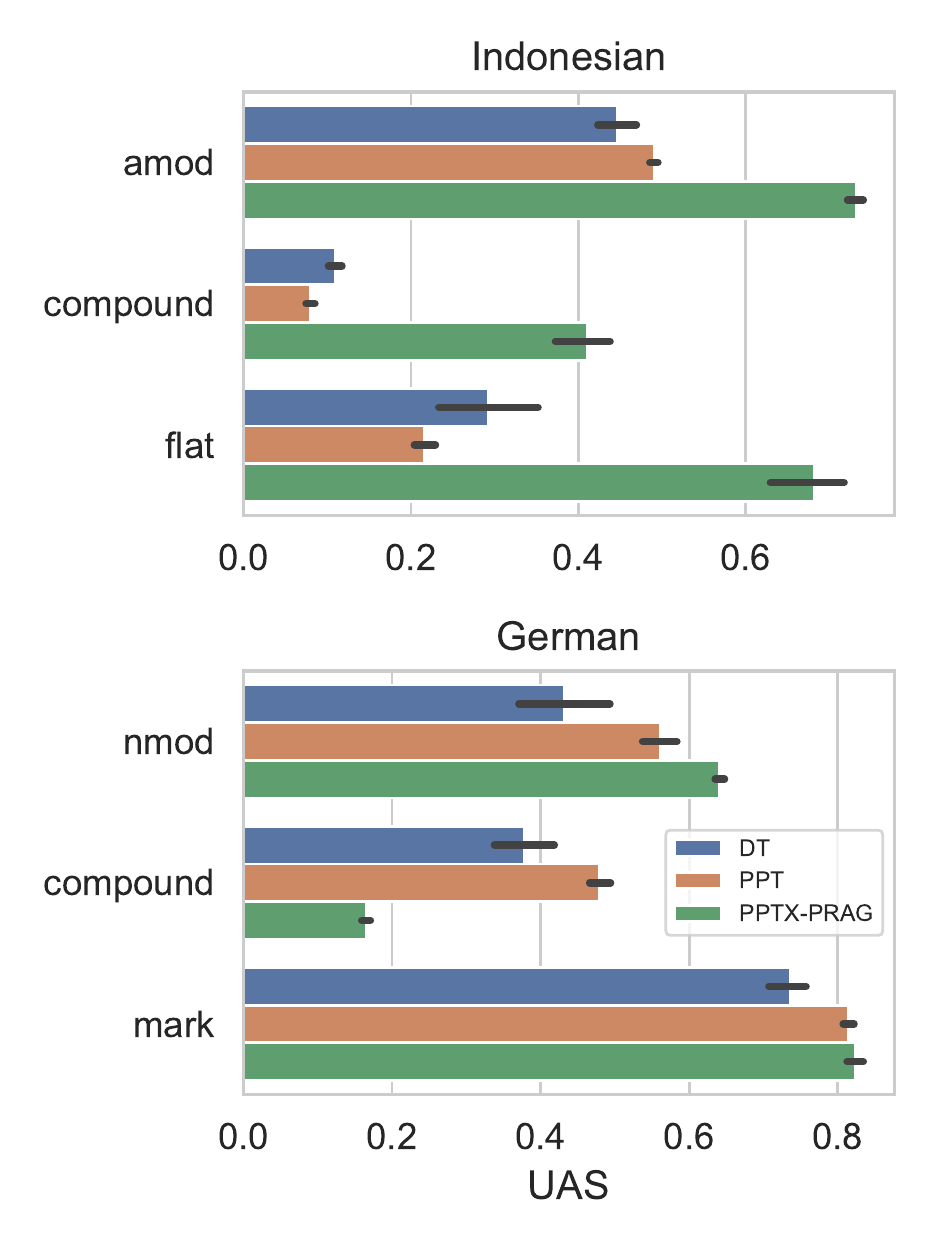}
  \caption{Comparison of direct transfer (DT), \AAST{}, and \AAETPRAG{} on select
    dependency labels of Indonesian (top) and German
    (bottom).}\label{fig:label-wise-eval}
\end{figure}

Next, we break down the performance of our methods based on the dependency
labels
to study their failure and success patterns. \cref{fig:label-wise-eval} shows
the UAS of DT, \AAST{}, and \AAETPRAG{} on Indonesian and German for select
dependency labels.

Looking at Indonesian, \AAST{} is slightly worse than DT in
terms of overall accuracy scores~(\cref{tbl:main}), and this is reflected across
dependency labels. However, we see in \cref{fig:label-wise-eval} that \AAST{}
outperforms DT on \texttt{amod}. In Indonesian,
adjectives follow the noun they modify, while in English the opposite is true in
general. Thus,
unsupervised target language adaptation seems able to address these kinds of
discrepancy between the source and target language. We find that \AAETPRAG{}
outperforms both DT and \AAST{} across dependency labels, especially on
\texttt{flat} and
\texttt{compound} labels as shown in \cref{fig:label-wise-eval}. Both labels
are related
to multi-word expressions (MWEs), so \AAET{} appears to improve parsing MWEs
in
Indonesian significantly.

For German we find that both \AAST{} and \AAETPRAG{} outperform DT on most
dependency labels, with the most notable gain on \texttt{nmod}, which appear in
diverse, and often non-local relations in both languages many of which do not
structurally translate, and fine-tuning improves performance as expected. Also,
we see \AAETPRAG{} significantly underperforms on \texttt{compound} while
\AAST{} is better than DT. German compounds are often merged into a single
token, and self-training appears to alleviate over-prediction of such relations.
The multi-source case may contain too much diffuse signal on \texttt{compound}
and thus the performance is worse than that of DT. We find that \AAST{} and
\AAET{} improves over DT on \texttt{mark}, likely because markers are often used
in places where German deviates from English by becoming verb-final (e.g.,
subordinate clauses). Both \AAST{} and \AAETPRAG{} seem able to learn this
characteristic as shown by their performance improvements. This analysis
suggests that the benefits of self-training depend on the syntactic properties
of the target language.

\subsection{Effect of Projectivity}

\begin{table}\small
  \centering
  \begin{tabular}{@{}l*{5}{S}@{}}
    \toprule
    \multirow{2}{*}{Model} & \multicolumn{4}{c}{Target} & {\multirow{2}{*}{AVG}} \\
    \cmidrule(lr){2-5}
                & {id} & {hr} & {fr} & {nl} &      \\
    \midrule
    \multicolumn{6}{c}{\textit{Non-projective}}    \\
    \addlinespace
    DT          & 45.2 & 51.9 & 68.1 & 59.5 & 56.2 \\
    \AAST{}     & 44.7 & 54.2 & 72.8 & 61.5 & 58.3 \\
    \AAETPRAG{} & 57.4 & 62.2 & 77.9 & 66.4 & 66.0 \\
    \midrule
    \multicolumn{6}{c}{\textit{Projective}}        \\
    \addlinespace
    DT          & 45.7 & 52.1 & 68.4 & 59.6 & 56.4 \\
    \AAST{}     & 45.0 & 54.0 & 72.3 & 61.7 & 58.3 \\
    \AAETPRAG{} & 57.5 & 61.1 & 78.1 & 67.7 & 66.1 \\
    \bottomrule
  \end{tabular}
  \caption{Comparison of projective and non-projective direct transfer (DT),
    \AAST{}, and \AAETPRAG{}. Scores are LAS, averaged over 5
    runs.}\label{tbl:proj-vs-nonproj}
\end{table}

In this experiment, we study the effect of projectivity on the performance of
our methods. We emulate a projective parser by restricting
the trees in $\tilde{Y}(x)$ to be projective. In other words, the sum in
\cref{eqn:aa-loss} is performed only over projective trees. At test time, we
search for the highest scoring projective tree. We compare DT, \AAST{}, and
\AAETPRAG{}, and report LAS on Indonesian (id) and Croatian (hr) as distant
languages, and on French (fr) and Dutch (nl) as nearby languages. The trend for
UAS and on the other languages is similar. We use the dynamic programming
implementation provided by \texttt{torch-struct} for the projective
case~\citep{rush2020}. We find that it consumes more memory than our Matrix-Tree
Theorem implementation, so we set the length cutoff to 20
tokens.\footnote{Hyperparameters are tuned; values are shown in
  \cref{tbl:main-hyper}.}

\cref{tbl:proj-vs-nonproj} shows result of our experiment, which suggests that
there is no significant performance difference between the projective and
non-projective variant of our methods. This result suggests that our methods
generalise well to both projective and non-projective parsing. That said, we
recommend the non-projective variant as it allows better parsing of languages that
are predominantly non-projective. Also, we find that it runs roughly 2x
faster than the projective variant in practice.

\subsection{Disentangling the Effect of Ensembling and Larger Data Size}

\begin{table}
  \centering
  \sisetup{
    table-format = 2.1,
    table-auto-round
  }
  \begin{tabular}{@{}lSS@{}}
    \toprule
    {\multirow{2}{*}{Model}} & \multicolumn{2}{c}{Target} \\
    \cmidrule(l){2-3}
                  & {ar}  & {es}  \\
    \midrule
    DT            & 28.09 & 64.11 \\
    \AAST{}       & 30.84 & 67.27 \\
    \AAETen{}     & 30.92 & 66.25 \\
    \AAETPRAGsm{} & 36.46 & 70.32 \\
    \AAETPRAG{}   & 36.45 & 71.88 \\
    \bottomrule
  \end{tabular}
  \caption{Comparison of LAS on Arabic and Spanish on the development set,
    averaged over 5 runs. \AAETen{} is \AAET{} with 5 English parsers as
    source, each trained on 1/5 size of the English corpus. \AAETPRAGsm{} is
    \AAET{} with the pragmatic selection of source languages (\AAETPRAG{}) but each
    source parser is trained on the same amount of data as
    \AAETen{}.}\label{tbl:disent}
\end{table}

The effectiveness of \AAET{} can be attributed to at least three factors: (1)~the
effect of ensembling source parsers (\textit{ensembling}), (2)~the effect of
larger data size used for training the source parsers (\textit{data}), and
(3)~the diversity of syntactic biases from multiple source languages
(\textit{multilinguality}). In this experiment, we investigate to what extent
each of those factors contributes to the overall performance. To this end, we
design two additional comparisons: \AAETen{} and \AAETPRAGsm{}.

\AAETen{} is \AAET{} with only English source parsers, where each parser
is trained on 1/5 of the English training set. That is, we randomly split
the English training set into five equal-sized parts, and
train a separate
parser on each. These parsers then serve as the source parsers for
\AAETen{}. Thus, \AAETen{} has the benefit of ensembling but not data and
multilinguality compared with \AAST{}.

\AAETPRAGsm{} is \AAET{} whose source language selection is the same as \AAETPRAG{},
but each source parser is trained on the training data whose size is roughly the same
as that of the training data of \AAETen{} source parsers. In other words, the
training data size is roughly equal to 1/5 of the English training set. To
obtain this data, we randomly sub-sample the training data of each source
language to the appropriate number of sentences. Therefore,  \AAETPRAGsm{} has
the benefit of
ensembling and multilinguality but not data.

\cref{tbl:disent} reports their LAS on the development set of Arabic and
Spanish, averaged over five runs. We also include the results of \AAETPRAG{} that
enjoys all three benefits. We observe that \AAST{} and \AAETen{} perform similarly
on Arabic, and \AAETen{} has a slightly lower performance on Spanish.
This result suggests a negligable effect of ensembling on
performance. On the other hand, \AAETPRAGsm{} outperforms \AAETen{}
remarkably, with approximately \SI{6}{\percent} and \SI{4}{\percent} LAS
improvement on Arabic and Spanish respectively, showing that
multilinguality has a much larger effect on performance than
ensembling.
Lastly, we see that \AAETPRAG{} performs similarly to
\AAETPRAGsm{} on Arabic,
and about \SI{1.6}{\percent} better on Spanish. This result
demonstrates that data size has an effect, albeit a smaller one compared to
multilinguality. To conclude, the effectiveness of \AAET{} can be attributed to
the diversity contributed through multiple languages, and not to ensembling
or larger source data
sets.

\section{Related Work}

Cross-lingual dependency parsing has been extensively studied in NLP. The
approaches can be grouped into two main categories. On the one hand, there are
approaches that operate on the data level. Examples of this category include
annotation projection, which aims to project dependency trees from a source
language to a target language~\citep{hwa2005,li2014b,lacroix2016,zhang2019d};
and source treebank reordering, which manipulates the
source language treebank to obtain another treebank whose statistics
approximately match those of the target language~\citep{wang2018a,rasooli2019}.
Both methods have no restriction on the type of parsers as they are
only concerned with the data. Transferring from multiple source languages with
annotation projection is also feasible~\citep{agic2016}.

Despite their effectiveness, these data-level methods may require access to the
source language data, hence are unusable when it is inaccessible due to privacy
or legal reasons. In such source-free transfer, only a model pre-trained on the
source language data is available. By leveraging parallel data, annotation
projection is indeed feasible without access to the source language data. That
said, parallel data is limited for low-resource languages or may have a poor
domain match. Additionally, these methods involve training the parser from
scratch for every new target language, which may be prohibitive.

On the other hand, there are methods that operate on the model
level. A typical approach is direct transfer (aka.,~zero-shot
transfer) which trains a parser on source language data, and then directly uses
it to
parse a target language. This approach is enabled by the shared input
representation between the source and target language such as POS
tags~\citep{zeman2008} or cross-lingual embeddings~\citep{guo2015,ahmad2019}.
Direct transfer supports source-free transfer and only requires training a
parser once on the source language data. In other words, direct transfer is
unsupervised as far as target language resources.

Previous work has shown that unsupervised target language adaptation outperforms
direct transfer. Recent work by \citet{he2019a} used a neural
lexicalised dependency model with valence (DMV)~\citep{klein2004} as the source
parser and fine-tuned it in an unsupervised manner on the unlabelled target
language data. This adaptation method allows for source-free transfer and
performs especially well on distant target languages. A different approach is
proposed by \citet{meng2019}, who gathered target language corpus statistics to
derive constraints to guide inference using the source parser. Thus, this
technique also allows for source-free transfer. A different method is proposed
by \citet{ahmad2019a} who explored the use of unlabelled data from an auxiliary
language, which can be different from the target language. They employed
adversarial training to learn language-agnostic representations. Unlike the
others, this method can be extended to support multi-source transfer. An older
method is introduced by \citet{tackstrom2013}, who leveraged ambiguity-aware
training to achieve unsupervised target language adaptation.
Their method is usable for both source-free and multi-source transfer. However,
to the best of our knowledge, its use for neural dependency parsing has not been
investigated. Our work extends theirs by employing it for the said purpose.

The methods of both \citet{he2019a} and \citet{ahmad2019a} have several
limitations. The method of \citet{he2019a} requires the parser to be generative
and projective.
Their generative parser is quite impoverished with an accuracy
that is \num{21} points lower than a state-of-the-art discriminative
arc-factored parser on
English. Thus, their choice of generative parser may constrain its potential
performance. Furthermore, their method performs substantially worse than direct
transfer on nearby target languages. Because of the availability of resources
such as Universal Dependency Treebanks~\citep{nivre2018}, it is likely that a
target language has a closely related high-resource language which can serve as
the source language. Therefore, performing well on nearby languages is more
desirable pragmatically. On top of that, it is unclear how to employ this method
for multi-source transfer. The adversarial training method of \citet{ahmad2019a}
does not suffer from the aforementioned limitations but is unusable for
source-free transfer. That is, it assumes access to the source language data,
which may not always be feasible due to privacy or legal reasons.

\section{Conclusions}

This paper presents a set of effective, flexible, and conceptually
simple methods for unsupervised cross-lingual dependency parsing, which can
leverage the power of state-of-the-art pre-trained neural network parsers. Our
methods improve over direct transfer and strong recent unsupervised transfer
models, by using source parser uncertainty for implicit supervision,
leveraging only unlabelled data in the target language. Our experiments show
that the
methods are effective for both single-source and multi-source transfer, free
from the limitations of recent transfer models, and perform well for
non-projective parsing. Our analysis shows that the effectiveness of the
multi-source transfer method is attributable to its ability to leverage diverse
syntactic signals from source parsers from different languages. Our findings
motivate future research into advanced methods for generating informative sets
of candidate trees given one or more source parsers.

\section*{Acknowledgments}

We thank the anonymous reviewers for the useful feedback. A graduate research
scholarship is provided by Melbourne School of Engineering to Kemal Kurniawan.

\bibliography{xduft,ud}
\bibliographystyle{acl_natbib}

\appendix

\section{Hyperparameter values}

Here we report the hyperparameter values for experiments presented in the paper.
\cref{tbl:src-hyper} shows the hyperparameter values of our English source
parser explained in \cref{sec:setup}. \cref{tbl:main-hyper} reports the
tuned hyperparameter values for our experiments shown in \cref{tbl:main},
\cref{fig:src-select}, and \cref{tbl:proj-vs-nonproj}.

\begin{table}[!h]\small
  \centering
  \begin{tabular}{@{}lr@{}}
    \toprule
    Hyperparameter                & Value     \\
    \midrule
    Sentence length cutoff        & 100       \\
    Word embedding size           & 300       \\
    POS tag embedding size        & 50        \\
    Number of attention heads     & 10        \\
    Number of Transformer layers  & 6         \\
    Feedforward layer hidden size & 512       \\
    Attention key vector size     & 64        \\
    Attention value vector size   & 64        \\
    Dropout                       & 0.2       \\
    Dependency arc vector size    & 512       \\
    Dependency label vector size  & 128       \\
    Batch size                    & 80        \\
    Learning rate                 & \num{e-4} \\
    Early stopping patience       & 50        \\
    \bottomrule
  \end{tabular}
  \caption{Hyperparameter values of the source parser.}\label{tbl:src-hyper}
\end{table}

\begin{table}\small
  \centering
  \begin{tabular}{@{}lrr@{}}
    \toprule
    \multirow{2}{*}{Hyperparameter} & \multicolumn{2}{c}{Value}   \\
                                    & Nearby       & Distant      \\
    \midrule
    \multicolumn{3}{c}{{ST}}                                      \\
    \addlinespace
    Sentence length cutoff          & 60           & 60           \\
    Learning rate                   & \num{5.6e-4} & \num{3.7e-4} \\
    L2 coefficient ($\lambda$)      & \num{3e-4}   & \num{2.8e-4} \\
    \midrule
    \multicolumn{3}{c}{{\AAST{}}}                                 \\
    \addlinespace
    Learning rate                   & \num{3.8e-5} & \num{2e-5}   \\
    L2 coefficient ($\lambda$)      & \num{0.01}   & \num{0.39}   \\
    \midrule
    \multicolumn{3}{c}{\AAET{}/\AAETLOO{}}                        \\
    \addlinespace
    Learning rate                   & \num{2.1e-5} & \num{5.9e-5} \\
    L2 coefficient ($\lambda$)      & \num{0.079}  & \num{1.2e-4} \\
    \midrule
    \multicolumn{3}{c}{{\AAETREPR{}}}                             \\
    \addlinespace
    Learning rate                   & \num{1.7e-5} & \num{9.7e-5} \\
    L2 coefficient ($\lambda$)      & \num{4e-4}   & \num{0.084}  \\
    \midrule
    \multicolumn{3}{c}{{\AAETPRAG{}}}                             \\
    \addlinespace
    Learning rate                   & \num{4.4e-5} & \num{8.5e-5} \\
    L2 coefficient ($\lambda$)      & \num{2.7e-4} & \num{2.8e-5} \\
    \midrule
    \multicolumn{3}{c}{Projective \AAST{}}                        \\
    \addlinespace
    Sentence length cutoff          & 20           & 20           \\
    Learning rate                   & \num{e-4}    & \num{e-4}    \\
    L2 coefficient ($\lambda$)      & \num{7.9e-4} & \num{7.9e-4} \\
    \midrule
    \multicolumn{3}{c}{Projective \AAETPRAG{}}                    \\
    \addlinespace
    Sentence length cutoff          & 20           & 20           \\
    Learning rate                   & \num{9.4e-5} & \num{9.4e-5} \\
    L2 coefficient ($\lambda$)      & \num{2.4e-4} & \num{2.4e-4} \\
    \bottomrule
  \end{tabular}
  \caption{Hyperparameter values of ST, \AAST{}, \AAET{}, \AAETREPR{},
    \AAETPRAG{}, projective \AAST{}, and projective
    \AAETPRAG{}. Sentence length cutoff for \AAST{}, \AAET{}, \AAETREPR{}, and
    \AAETPRAG{} is 30, as explained in \cref{sec:setup}.}\label{tbl:main-hyper}
\end{table}

\end{document}